\documentclass[10pt,twocolumn,letterpaper]{article}

\usepackage[final]{cvpr}              % To produce the CAMERA-READY version
\usepackage{multirow}
\usepackage{booktabs}

\usepackage[dvipsnames]{xcolor}

\definecolor{cvprblue}{rgb}{0.21,0.49,0.74}

\definecolor{cvprblue}{rgb}{0.21,0.49,0.74}
\usepackage[pagebackref,breaklinks,colorlinks,citecolor=cvprblue]{hyperref}

\def\paperID{11144} % *** Enter the Paper ID here
\def\confName{CVPR}
\def\confYear{2024}

\title{Rethinking Prior Information Generation with CLIP for Few-Shot Segmentation}

\author{Jin Wang$^{1}$ \hspace{12pt}Bingfeng Zhang$^1$\thanks{Corresponding author.} \hspace{12pt} Jian Pang$^{1}$ \hspace{12pt}Honglong Chen$^{1}$ \hspace{12pt} Weifeng Liu$^1$\footnotemark[1] \\
		$^1$China University of Petroleum (East China)\hspace{12pt}\\ 
		{\tt\small\{wangjin, jianpang\}@s.upc.edu.cn, \{bingfeng.zhang, wfliu, chenhl\}@upc.edu.cn}
}

\begin{document}
\maketitle
\begin{abstract}
Few-shot segmentation remains challenging due to the limitations of its labeling information for unseen classes. Most previous approaches rely on extracting high-level feature maps from the frozen visual encoder to compute the pixel-wise similarity as a key prior guidance for the decoder. However, such a prior representation suffers from coarse granularity and poor generalization to new classes since these high-level feature maps have obvious category bias. In this work, we propose to replace the visual prior representation with the visual-text alignment capacity to capture more reliable guidance and enhance the model generalization. Specifically, we design two kinds of training-free prior information generation strategy that attempts to utilize the semantic alignment capability of the Contrastive Language-Image Pre-training model (CLIP) to locate the target class. Besides, to acquire more accurate prior guidance, we build a high-order relationship of attention maps and utilize it to refine the initial prior information. Experiments on both the PASCAL-5{i} and COCO-20{i} datasets show that our method obtains a clearly substantial improvement and reaches the new state-of-the-art performance. The code is available on the project website~\footnote{https://github.com/vangjin/PI-CLIP}.
\end{abstract}

% The sourcode is available at \href{https://github.com/vangjin/PI-CLIP}{https://github.com/vangjin/PI-CLIP}.

% Besides, to acquire more representative and efficient features, we also expand the information of semantic features by combining the feature extraction capability of vision model and CLIP model.    
\section{Introduction}
\label{sec:intro}

With the development of deep learning~\cite{cui2023generalized, cheng2021per}, semantic segmentation~\cite{tian2023learning, long2015fully, minaee2021image, zhang2023coinseg, jiao2023learning} has made a great progress. Traditional semantic segmentation relies on intensive annotations which is time-consuming and labour-intensive, once the segmentation model encounters samples with limited labeled data, it cannot output accurate prediction, making it difficult to apply in practice.  Few-shot segmentation~\cite{lang2022learning, tian2020prior, zhang2021self} is proposed to address the above problem, which aims to segment novel classes with a few annotated samples during inference. To achieve this, it divides data into a support set and a query set, the images in the query set are segmented using the provided information in the support set as a reference.
\begin{figure}[!t]
	\centering
	\includegraphics[width=\columnwidth]{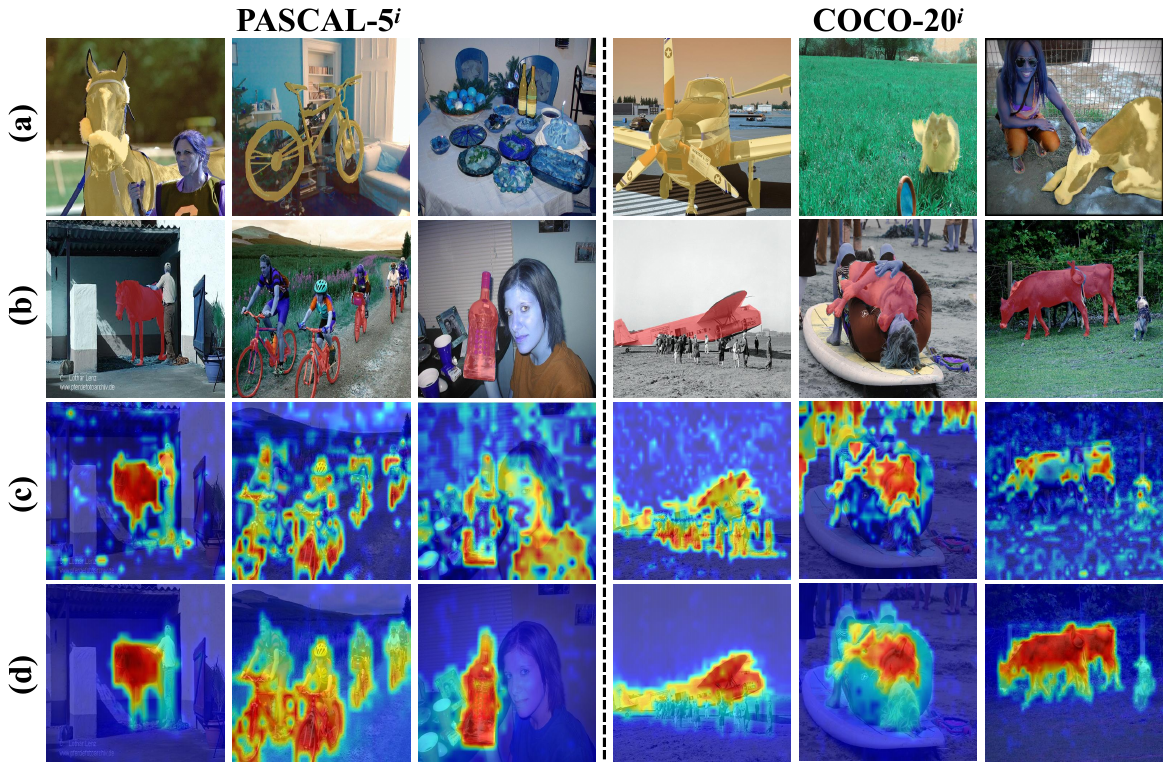}
        \vspace{-1.5em}
	\caption{Comparison of prior information. (a) Support images with ground-truth masks; (b) Query images with ground-truth masks; (c) Prior information from previous approaches generated based on the frozen ImageNet~\cite{deng2009imagenet} weights, which are biased towards some classes, such as the `Person' class; (d) Our prior information, which is generated utilizing the text and visual alignment ability of the frozen CLIP model. Our prior information is finer-grained and mitigates the bias of the class.}
 % Comparison of prior masks. Previous approaches generate prior masks (the 3rd row) by directly interacting with high-dimensional features from frozen ImageNet~\cite{deng2009imagenet} weights, which are biased towards some classes, such as the `Person' class. We propose to utilize the frozen CLIP model to generate finer-grained masks (the 4th row) and mitigate the bias of the class.
	\label{fig:1}
\end{figure}
Existing few-shot segmentation methods can be roughly categorized into two types: pixel-level matching~\cite{zhang2019pyramid, yang2020brinet,yang2020brinet, hong2022cost, shi2022dense} and prototype-level matching~\cite{tian2020prior, lang2022learning, peng2023hierarchical, fan2022self, zhang2021self, wang2023rethinking}. Pixel-level matching uses the pixel-to-pixel matching mechanism~\cite{rakelly2018few, zhang2022mfnet, shi2022dense} to enforce the few-shot model mine pixel-wise relationship~\cite{bi2023not, tavera2022pixel, zhang2022mfnet}. Prototype-level matching methods extract prototypes~\cite{wang2019panet, liu2023multi, huang2023prototypical, okazawa2022interclass} from the support set to perform similarity~\cite{lee2022pixel, peng2023hierarchical, he2023prototype, pandey2022adversarially} or dense comparisons~\cite{chen2021apanet, li2021adaptive, ding2023self, lang2022beyond, sun2023attentional} with the query image features to make predictions. No matter pixel-level matching or prototype-level matching, most recent approaches~\cite{peng2023hierarchical,lang2022learning, fan2022self, li2021adaptive, siam2019adaptive} introduce the prior masks~\cite{lang2022learning, zhang2021rich, liu2022dynamic} as a coarse localization map to guide the matching or segmentation process to concentrate on the located regions. However, such prior masks are mainly generated through interacting fixed high-dimensional features from the visual pre-trained models, \ie, CNN with ImageNet~\cite{deng2009imagenet} pre-train initialization, causing several insolvable problems as shown in Fig.~\ref{fig:1}: 1) incorrect target location response due to original ImageNet~\cite{deng2009imagenet} pre-training weights being insensitive to category information, which misleads the segmentation process and thus restricting generalization of the model. 2) coarse prior mask shapes, caused by undistinguished vision features between the target and non-target pixels, make the prior information locate many non-target regions, which further confuses the segmentation process.

% The current approaches mainly interacts fixed high-dimensional features from the visual pre-trained models, \eg, CNN with ImageNet~\cite{deng2009imagenet} pre-train initialization, to generate the prior masks. However, as shown in Fig.~\ref{fig:1}, these fixed high-dimensional features are not sensitive enough to category information, leading to the following problems: 1) coarse prior mask shapes, which caused by undistinguished vision features between foreground and background, making the feature similarity cannot generate accurate object regions. 2) incorrect category response due to original Imagenet~\cite{deng2009imagenet} pre-training weights being insensitive to category information, making the prior mask locate wrong target object. These two problems further limit the generalization of the few-shot segmentation task.
% The recently emerged CLIP model has been successful in few-shot tasks
To address the aforementioned drawbacks, we rethink the prior mask generation strategy and attempt to use Contrastive Language-Image Pre-training (CLIP)~\cite{radford2021learning} to generate more reliable prior information for few-shot segmentation. A large amount of text-image training data pairs make the CLIP model sensitive to category due to the forced text-image alignment, which enables better localization of the target class~\cite{luddecke2022image, zhu2023not, zhang2022tip}. Besides, the success in the zero-shot task~\cite{lin2023clip, radford2021learning, guo2023calip} also demonstrates the powerful generalization ability of the CLIP model. Based on this, we attempt to utilize the CLIP model to generate better prior guidance.

Finally, in this paper, we propose Prior Information Generation with CLIP (PI-CLIP), a training-free CLIP-based approach, to extract prior information to guide the few-shot segmentation. Specifically, we propose two kinds of prior information generation, the first one is called visual-text prior information (VTP) which aims to provide accurate prior location based on the strong visual-text alignment ability of the CLIP model, we re-design target and non-target prompts and force the model to perform category selection for each pixel, thus locating more accurate target regions. The other one is called visual-visual prior information (VVP) which focuses on providing more general prior guidance using the matching map extracted from the CLIP model between the support set and the query image. 

% However, as a training-free approach, VTP excessively focuses on local regions and VVP adds many non-target pixels in the prior information, both of which cannot provide finer-grained prior guidance. Therefore, we build a high-order attention matrix based on the attention maps of the CLIP model, called Prior Information Refinement (PIR),  to refine the initial VTP and VVP, which makes full use of the pixel-pair relationship to highlight the target area and reduce the response to the non-target area, thus clearly improving the quality of the prior mask. Without any training, the generated prior masks overcome the drawback caused by inaccurate prior information in existing methods, significantly improving the performance of different few-shot approaches.
However, as a training-free approach, the forced alignment of visual information and text information makes VTP excessively focus on local target regions instead of the expected whole target regions, the incomplete original global structure information only highlights local target regions which reduces the quality of guidance. Based on this, we build a high-order attention matrix based on the attention maps of the CLIP model, called Prior Information Refinement (PIR), to refine the initial VTP, which makes full use of the original pixel-pair structure relationship to highlight the whole target area and reduce the response to the non-target area, thus clearly improving the quality of the prior mask. Note that VVP is not refined to keep its generalization ability. Without any training, the generated prior masks overcome the drawback caused by inaccurate prior information in existing methods, significantly improving the performance of different few-shot approaches.
%Extensive experiments show that our approach outperforms other approaches by a clear margin, reaching new state-of-the-art performance.
% Specifically, in visual-text prior information generation, we define foreground prompt and background prompt for text embedding and force the model to perform category selection for each pixel by using Gradient-weighted Class Activation Mapping (Grad-CAM)~\cite{zhou2016learning} which can accurately locate specific categories. In visual-visual prior information generation, in order to make full use of the support information and get a more generalized prior information, we calculate the visual similarity between clip support features and clip query features as a complement. 

Our contributions are as follows:
\begin{itemize}
 	\item We rethink the prior information generation for few-shot segmentation, proposing a training-free strategy based on the CLIP model to provide more accurate prior guidance by mining visual-text alignment information and visual-visual matching information. 
    \item To generate finer-grained prior information, we build a high-order attention matrix to refine the initial prior information based on the frozen CLIP attention maps to extract the relationship of different pixels, clearly improving the quality of the prior information.
    \item Our method has a significant improvement over existing methods on both PASCAL-5$^{i}$~\cite{shaban2017one} and COCO-20$^{i}$~\cite{nguyen2019feature} datasets and achieves state-of-the-art performance.
\end{itemize}

% As shown in Fig.1(a), although the existing method supplements the effective prototype information for guidance, the coarse prior mask leads to a poor segmentation result. If we use training data to train a network using a supervised approach to generate a more detailed prior mask, as shown in Fig.1(b), the model can be said to be completely dependent on the accurate prior mask for prediction to a certain extent and get a better performance, but once it encounters the problem of insufficient labeling data, for example, during the testing process, it does not have the ability to generalize, which leads to the poor quality of the generated prior mask, which in turn leads to the model performance collapse. Based on this, we rethink the mask generation method with the powerful semantic alignment ability of multi-modal models, and adopt the training-free strategy to generate prior masks, as can be seen from Fig.1(c), this method is not limited by the amount of labeled data, and can well guide the model for segmentation.
 \begin{figure*}[!t]
	\centering
	\includegraphics[width=0.95\textwidth]{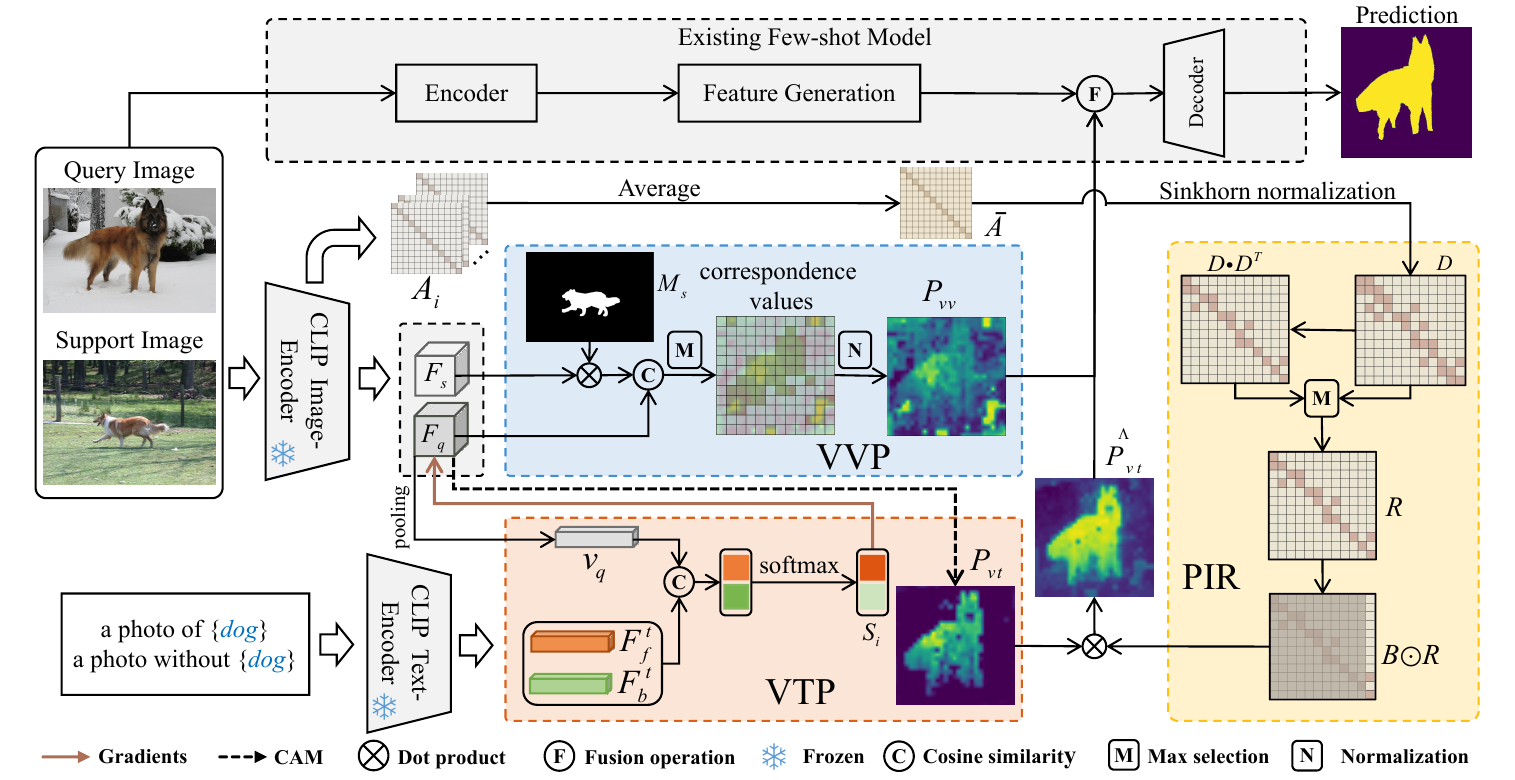}
	\caption{Overview of our proposed PI-CLIP for few-shot segmentation. We design a group of text prompts for a certain class to attract more attention to target regions. The VTP module generates the visual-text prior information by aligning the visual information and text information with the help of softmax-GradCAM. The VVP module generates the visual-visual prior information by a pixel-level similarity calculation. The PIR module is proposed to refine the coarse initial prior information. Finally, the original prior information in the existing few-shot model is directly replaced by VVP and refined VTP, after passing the decoder, the final prediction is generated.}
	\label{fig:2}
\end{figure*}

\section{Related Work}
\subsection{Few-Shot Segmentation} 
Few-shot segmentation aims to generate dense predictions for new classes using a small number of labeled samples. Most existing few-shot segmentation methods followed the idea of metric-based meta-learning~\cite{hu2019attention, sung2018learning}. Depending on the object of the metric, current approaches can be divided into pixel-level matching mechanism~\cite{yang2020brinet, hong2022cost, shi2022dense} and prototype-level matching mechanism~\cite{fan2022self, li2021adaptive, siam2019adaptive, zhang2021rich, liu2022dynamic}. No matter pixel-level matching or prototype-level matching mechanism, most recent approaches~\cite{tian2020prior, luo2021pfenet++, lang2022learning, yang2023mianet, zhang2021self} utilized prior information to guide the segmentation process.

 PCN~\cite{lu2021simpler} fused the scores from base and novel classifier to prevent base class bias. CWT~\cite{lu2023prediction} adapted the classifier's weights to each query image in an inductive way. PFENet~\cite{tian2020prior} first proposed to utilize prior information extracted from pixel relationship between support set and query image to guide the decoder and designed a module to aggregate contextual information at different scales. PFENet++~\cite{luo2021pfenet++} rethinked the prior information and proposed to utilize the additional nearby semantic cues for a better location ability of the prior information. BAM~\cite{lang2022learning} further optimized the prior information and proposed to leverage the segmentation of new classes by suppressing the base classes learned by the model. SCL~\cite{zhang2021self} proposed a self-guided learning approach to mine the lost critical information on the prototype and utilize the prior information as guidance for the decoder. IPMT~\cite{liu2022intermediate} mined useful information by interacting prototype and mask to mitigate the category bias and design an intermediate prototype to mine more accurate prior guidance by an iterative approach. 
 MM-Former~\cite{zhang2022mask} utilized a class-specific segmenter to decompose the query image into a single possible prediction and extracted support information as prior to matching the single prediction which can improve the flexibility of the segmentation network. MIANet~\cite{yang2023mianet} proposed to use general prior information from semantic word embedding and instance information to perform an accurate segmentation. HDMNet~\cite{peng2023hierarchical} mined pixel-level correlation with transformer based on two kinds of prior information between support set and query image to avoid overfitting.
 
 Most recent existing methods utilized coarse masks to guide segmentation, our approach attempts to generate finer-grained masks with the help of CLIP models.
 
\subsection{Contrastive Language-Image Pretraining}
 Contrastive Language-Image Pretraining (CLIP)~\cite{radford2021learning} is able to map text and image into high-dimensional space by text-encoder and image-encoder respectively. Trained on a large amount of text-image data makes the CLIP~\cite{radford2021learning, huang2023clip2point} model has a strong feature extraction capability, which is used in many downstream applications such as detection~\cite{ju2022adaptive}, segmentation~\cite{lin2023clip, yang2023multi, shuai2023visual}, and so on. CLIPSeg~\cite{luddecke2022image} first attempted to introduce the CLIP model into few-shot segmentation. However, CLIPseg is more like to use the CLIP model as a validation method to show the powerful capability of the CLIP model in few-shot tasks. In this paper, we design a new prior information generation strategy using the CLIP model for few-shot segmentation through the visual-text relationship and the visual-visual relationship to perform a more efficient guidance.

 \section{Method}
 \subsection{Task Description}
Few-shot segmentation aims to segment novel classes by using the model trained on base classes. Most existing few-shot segmentation approaches follow the meta-learning paradigm. The model is optimized with multiple meta-learning tasks in the training phase and evaluates the performance of the model in the testing phase. Given a dataset $D$, dividing it into a training set ${D_{train}}$ and a test set ${D_{test}}$, there has no crossover between the class set ${C_{train}}$ in the training set and class set ${C_{test}}$ in the test set (${C_{train}} \cap {C_{test}}= \emptyset$). The model is expected to transfer the knowledge in ${D_{train}}$ with restricted labeled data to the ${D_{test}}$. Both training set ${D_{train}}$ and test set ${D_{test}}$ are composed of support set $S$ and query set $Q$, support set $S$ contains $K$ samples $S=\{S_1,S_2,\,\ldots,S_K\}$, each $S_i$ contains an image-mask pair $\{I_s, M_s\}$ and query set $Q$ contains $N$ samples $Q=\{Q_1,Q_2,\,\ldots,Q_N\}$, each $Q_i$ contains an image-mask pair $\{I_q, M_q\}$. During training, the few-shot model is optimized with training set ${D_{train}}$ by epochs where the model performs prediction for query image $I_q$ with the guidance of the support set $S$. During inference, the performance will be acquired with the test set ${D_{test}}$, and the model is no longer optimized.

\subsection{Method Overview}
In order to enhance the ability of prior information to localize target categories as well as to produce more generalized prior information, we propose to mine visual-text and visual-visual information instead of purely visual feature similarity to guide the segmentation process. Besides, to further improve the quality of the prior information to get finer-grained guidance, we design an attention map-based high-order matrix to refine the initial prior information by pixel-pairs relationships, Fig.~\ref{fig:2} shows our framework of the one-shot case with the following steps:

\begin{enumerate}
\item Given a support image and a query image with the target class name, we first input the query image and support image to the CLIP image encoder to generate corresponding visual support and query features. Meanwhile, the target class name is used to build two text prompts, \ie, target prompt and non-target prompt, which are then input to the CLIP text encoder to generate two text embeddings.  
% 2) Then, to generate the initial visual-text prior information, we compute the classification scores using the distance between two text embeddings and the visual query features. After that, we use GradCAM to generate the response map for the target class as the visual-text prior information.
\item Then, two text embeddings and the query visual features are input to the visual-text prior (VTP) module to generate the initial VTP information by enforcing a classification process for each pixel.
% \item Meanwhile, the support visual features and query visual features are input to the visual-visual prior (VVP) module where the initial VVP information is generated through the pixel-level relationship. 
\item Meanwhile, the support visual features and query visual features are input to the visual-visual prior (VVP) module where the VVP information is generated through the pixel-level relationship.

% \item After that, we extract attention maps from the clip model, which are input to our prior information refinement (PIR) module to build a high-order attention matrix for refining the above initial VTP and VVP information.
\item After that, we extract attention maps from the clip model, which are input to our prior information refinement (PIR) module to build a high-order attention matrix for refining the above initial VTP information.

\item Finally, the original prior information in the existing method is directly replaced by our VVP and refined VTP to generate the final prediction for the query image.
\end{enumerate}

\subsection{Visual-Text Prior Information Generation}
Few-shot Segmentation (FSS) remains one major challenge that an image might have more than one class, but the model is required to segment only one class at each episode. This challenge means that once the prior information is unable to provide the correct target region, \eg, a true target region is ``dog" but the prior information provides a ``cat" region, it will confuse the FSS model to segment the true target pixels, especially for the untrained novel class. 
% Besides, although informing the model target class, the model encounters a new problem is that the existence of the real background area in addition to the target and non-target classes affects the location ability to the foreground regions of the target class.
To correctly locate target regions, we utilize the visual-text alignment information from the CLIP model to produce a new prior information called VTP. We innovatively define a group of text prompts of the target class as a guidance to the model, in which the target (foreground) text prompts $t_{f}$ is defined as ``a photo of \{\emph{target class}\}" and the non-target (background) text prompts $t_{b}$ is ``a photo without \{\emph{target class}\}".
% \emph{f, b} are the abbreviation for foreground and background which represents target and non-target respectively.

Based on the designed text prompts, a pixel-level classification is performed for the query image so as to locate the true target foreground regions. To force the model to decide whether one pixel is the target or not, we use softmax-GradCAM~\cite{lin2023clip} to generate the prior information using the relationship between the visual and text features. Specifically, the designed target and non-target prompts, \ie, ``a photo of \{\emph{target class}\}" and ``a photo without \{\emph{target class}\}", are sent to the CLIP text encoder to get the high dimensional text features, represented as $F_{f}^{t}$ and $F_{b}^{t}$. Suppose the query image is $I_q$, after passing the CLIP visual encoder, the query features $F_q^v \in \mathbb{R}^{d \times (hw+1)}$, after removing the class token in $F_q^v$, visual query feature $F_q \in \mathbb{R}^{d \times hw}$ is generated, then the query token $v_q$ is obtained through global average pooling:
\begin{equation}
\begin{aligned}
v_q = \frac{1}{hw} \sum_{i=1}^{hw} F_{q}(i), v_q \in \mathbb{R}^{d \times 1}.
\end{aligned}
\end{equation}

Then classification scores are obtained by performing a distance calculation between the text features and the query token after the softmax operation:
\begin{equation}
\begin{aligned}
S_i = softmax(\frac{v_q^\mathrm{T}F^t_i}{\|v_q\|\|F^t_i\|}/\tau), i \in \{ {f, b} \}
\end{aligned},
\end{equation}
where \emph{T} represents the matrix transposition and $\tau$ is a temperature parameter. Then the gradient is calculated based on the final classification score:
\begin{equation}
\begin{aligned}
w_{m} = \frac{1}{hw}\sum_{i}^{} \sum_{j}^{} \frac{\partial S_f }{\partial F_q^m(i, j) }
\end{aligned},
\end{equation}
where $w_{m}$ is the weight for $m$-$th$ feature map of the foreground regions, $F_q^{m}$ means the activation value for $m$-$th$ feature map and $(i, j)$ means the pixel position.

Finally, the visual-text prior information $P_{vt} \in \mathbb{R}^{1 \times h \times w}$ is obtained:
\begin{equation}
\begin{aligned}
P_{vt} = ReLU(\sum_{m}^{} w_m F_{q}^m )
\end{aligned},
\end{equation}
$ReLU$ means the ReLU activation function to filter the negative response. Due to the forced alignment of the semantic information from the visual modal and text modal with softmax-GradCAM, the generated visual-text prior information clearly locates accurate target regions, which avoids the confusion of the segmentation process.

\subsection{Visual-Visual Prior Information Generation}
We enforce VTP to make a classification for each pixel so that it can locate the correct region. However, we observe that VTP tends to locate a discriminative local region, \eg, the ``head" region of a ``dog" rather than the whole region. To overcome this drawback, we attempt to take advantage of the support information that is naturally present in few-shot segmentation and get region-larger and location-rougher prior information to give more generalized guidance to the model.

We design VVP to mine more general target information by performing matching on the visual-visual relationship between the support image feature and the query image feature. Suppose the support image is $I_s$, after passing through the CLIP image encoder, its high dimensional image feature is generated and the visual support feature is $F_s^{v} \in \mathbb{R}^{d \times hw}$ (class token is removed). To get more target-focused support information, we extract the target information from the support image:
\begin{equation}
\begin{aligned}
F_s = F_s^v \odot M_s
\end{aligned},
\end{equation}
where $M_s$ represents the support mask, which is required to downsample to the same height and width as the feature map. Then we perform a cosine similarity calculation between all pixel pairs for $f_s^i \in F_s$ and $f_q^j \in F_q$ as:
\begin{equation}
\begin{aligned}
cos(f_s^i, f_q^j)=\frac{(f_s^i)^\mathrm{T}f_q^j}{\|f_s^i\|\|f_q^j\|}\quad i,j \in \{0,1,2,\,\ldots hw\}
\end{aligned}.
\end{equation}

For each pixel in $F_q$, the maximum similarity is selected from all pixels in the support feature as the correspondence value:
\begin{equation} \label{max value}
\begin{aligned}
P_{vv}(j) = \max_{i,j \in \{1,2,\ldots,hw\}} cos(f_s^i, f_q^j)
\end{aligned}.
\end{equation}

After computing all correspondence value by the above equation, prior information is generated, the values in $P_{vv}$ are normalized by a min-max normalization to generate the initial visual-visual prior information, $P_{vv} \in \mathbb{R}^{1 \times h \times w}$:
\begin{equation}
\begin{aligned}
P_{vv} =\frac{P_{vv}-min(P_{vv})}{max(P_{vv})-min(P_{vv})+\varepsilon}
\end{aligned},
\end{equation}
where $\varepsilon$ is set to $10^{-7}$. We utilize the feature from the CLIP model which contains more reliable semantic information to acquire the visual-visual prior information, thus matching support information with the query image, the model can provide more general location information as the prior guidance.

\subsection{Prior Information Refinement}
% The prior information above is generated by the visual and textual features extracted from the frozen CLIP weights. As a training-free method, the representation of these two prior information can not adaptively guide the model to perform an efficient segmentation. To generate finer-grained prior information, we propose a Prior Information Refinement (PIR) module to refine initial prior information. PIR builds a high-order matrix based on the attention map from the query image, which can accurately build the pixel-wise relationship and thus efficiently refine the VTP and VVP prior information. In this way, the refined prior information pays more attention to target regions and focuses less on non-target regions due to the higher-order relationships between pixel points better capturing spatial information and details of semantics.
The above prior information is generated by the visual and textual features extracted from the frozen CLIP weights. As a training-free method, the representation of the prior information can not adaptively guide the model to perform an efficient segmentation. To generate finer-grained prior information that focuses more target regions, we propose a Prior Information Refinement (PIR) module to refine the initial prior information. PIR builds a high-order matrix based on the attention map from the query image, which can accurately build the pixel-wise relationship and retain the original global structure information, thus efficiently capturing spatial information and details of semantics to refine the prior information. In this way, the refined prior information pays more attention to the whole target regions and focuses less on non-target regions.

Specifically, suppose $A_i \in \mathbb{R}^{hw\times hw}$ is the multi-head self-attention map generated from CLIP with the $i$-$th$ block, to acquire more accurate attention maps for each image, we first compute the average attention map by:
\begin{equation}
\begin{aligned}
\overline{A} = \frac{1}{l} \sum_{i=n-l}^{n} A_i
\end{aligned},
\end{equation}
where $l$ and $n$ are the block number of the vision transformer in CLIP and $l<n$. Based on the average attention map, in order to eliminate as much as possible the influence of the background region while preserving the intrinsic structural information, we design a high-order refinement matrix $R \in \mathbb{R}^{1 \times h \times w}$ follows:
\begin{equation}
\begin{aligned}
R = max(D,(D\cdot D^{T})), D=Sinkhorn(\overline{A})
\end{aligned},
\end{equation}
where $Sinkhorn$ means Sinkhorn normalization~\cite{sinkhorn1964relationship} to aligning data from rows and columns.
We then utilize the refinement matrix $R$ to refine the initial coarse prior information from VTP and VVP by:
\begin{equation}
\begin{aligned}
\hat{P}_i =  B \odot R \cdot P_{i} , \{ i \in {vt, vv} \}
\end{aligned},
\end{equation}
where $B$ is a box mask generated from the prior mask following~\cite{lin2023clip} and $\odot$ represents the Hadamard product. We experimentally found that only refining the visual-text prior $P_{vt}$ is enough since the refinement matrix will make $P_{vt}$ and $P_{vv}$ produce similar responses, which will damage the generalization of the model. Therefore, we select the refined text-visual prior and initial visual-visual prior, \ie, $\hat{P}_{vt}$ and ${P}_{vv}$, as the final prior information.
% Finally, we directly replace the prior information in existing methods with the concatenation of our refined visual-text prior information $\hat{P}_{vv}$ and visual-visual prior information $\hat{P}_{vt}$, to generate the final prediction.

Finally, we directly replace the prior information in existing methods with the concatenation of our visual-visual prior information ${P}_{vv}$ and refined visual-text prior information $\hat{P}_{vt}$, to generate the final prediction.

\begin{table*}[h]
\caption{Performance comparisons with mIoU (\%) as a metric on PASCAL-5$^{i}$, ``\textbf{ours}-PI-CLIP (PFENet)", ``\textbf{ours}-PI-CLIP (BAM)" and ``\textbf{ours}-PI-CLIP (HDMNet)" represent the baseline is PFENet~\cite{tian2020prior}, BAM~\cite{lang2022learning} and HDMNet~\cite{peng2023hierarchical} respectively.}
\label{tab:tab1}
\centering
\resizebox{\textwidth}{!}{$
\begin{tabular}{llccccc|ccccc}
\hline
\multirow{2}*{Method} &\multirow{2}*{Backbone} &\multicolumn{5}{c|}{1-shot} &\multicolumn{5}{c}{5-shot}\\ 
\cline{3-12}
& &Fold0 &Fold1 &Fold2 &Fold3 &Mean &Fold0 &Fold1 &Fold2 &Fold3 &Mean\\
\hline
SCL (CVPR'21)~\cite{zhang2021self} & resnet50 & 63.0 & 70.0 & 56.5 & 57.7 & 61.8 & 64.5 & 70.9 & 57.3 & 58.7 & 62.9 \\
SSP (ECCV'22)~\cite{fan2022self} & resnet50 & 60.5 & 67.8 & 66.4 & 51.0 & 61.4 & 67.5 & 72.3 & 75.2 & 62.1 & 69.3\\
DCAMA (ECCV'22)~\cite{shi2022dense} & resnet50 & 67.5 & 72.3 & 59.6 & 59.0 & 64.6 & 70.5 & 73.9 & 63.7 & 65.8 & 68.5\\
NERTNet (CVPR'22)~\cite{liu2022learning} & resnet50 & 65.4 & 72.3 & 59.4 & 59.8 & 64.2 & 66.2 & 72.8 & 61.7 & 62.2 & 65.7 \\
IPMT (NeurIPS'22)~\cite{liu2022intermediate} & resnet50 & 72.8 & 73.7 & 59.2 & 61.6 & 66.8 & 73.1 & 74.7 & 61.6 & 63.4 & 68.2 \\ 
ABCNet (CVPR'23)~\cite{wang2023rethinking} & resnet50 & 68.8 & 73.4 & 62.3 & 59.5 & 66.0 & 71.7 & 74.2 & 65.4 & 67.0 & 69.6 \\
MIANet (CVPR'23)~\cite{yang2023mianet} & resnet50 & 68.5 & 75.8 & 67.5 & 63.2 & 68.8 & 70.2 & 77.4 & 70.0 & 68.8 & 71.6 \\
MSI (ICCV'23)~\cite{moon2023msi} & resnet50 & 71.0 & 72.5 & 63.8 & 65.9 & 68.3 & 73.0 & 74.2 & 66.6 & 70.5 & 71.1 \\
\hline
PFENet (TPAMI'20)~\cite{tian2020prior} & resnet50 & 61.7 & 69.5 & 55.4 & 56.3 & 60.8 & 63.1 & 70.7 & 55.8 & 57.9 & 61.9 \\
BAM (CVPR'22)~\cite{lang2022learning} & resnet50 & 68.9 & 73.6 & 67.6 & 61.1 & 67.8 & 70.6 & 75.1 & 70.8 & 67.2 & 70.9 \\
HDMNet (CVPR'23)~\cite{peng2023hierarchical} & resnet50 & 71.0 & 75.4 & 68.9 & 62.1 & 69.4 & 71.3 & 76.2 & 71.3 & 68.5 & 71.8 \\
\hline
\textbf{ours}-PI-CLIP (PFENet) & resnet50 &67.4  &76.5  &71.3  &69.4  &71.2  &70.4  &78.2  &72.4  &70.2  &72.8  \\ 
\textbf{ours}-PI-CLIP (BAM) & resnet50 &72.4  &80.2  &71.6  &70.5  &73.7  &72.6  &80.6  &73.5  &72.0  &74.7  \\ 
\textbf{ours}-PI-CLIP (HDMNet) & resnet50 & \textbf{76.4} & \textbf{83.5} & \textbf{74.7} & \textbf{72.8} &\textbf{76.8} & \textbf{76.7} & \textbf{83.8} & \textbf{75.2} &\textbf{73.2} & \textbf{77.2} \\\hline
\end{tabular}%
$}
\end{table*}

\begin{table*}[!t]
\caption{Performance comparisons on COCO-20$^{i}$, ``\textbf{ours}-PI-CLIP (HDMNet)" represent the baseline is HDMNet~\cite{peng2023hierarchical}.}
\label{tab:tab2}
\centering
\resizebox{\textwidth}{!}{$
\begin{tabular}{llccccc|ccccc}
\hline
\multirow{2}*{Method} &\multirow{2}*{Backbone} &\multicolumn{5}{c|}{1-shot} &\multicolumn{5}{c}{5-shot}\\ 
\cline{3-12}
& &Fold0 &Fold1 &Fold2 &Fold3 &Mean &Fold0 &Fold1 &Fold2 &Fold3 &Mean\\
\hline
SCL (CVPR'21)~\cite{zhang2021self} & resnet50 & 36.4 & 38.6 & 37.5 & 35.4 & 37.0 & 38.9 & 40.5 & 41.5 & 38.7 & 39.9 \\
SSP (ECCV'22)~\cite{fan2022self} & resnet101 & 39.1 & 45.1 & 42.7 & 41.2 & 42.0 & 47.4 & 54.5 & 50.4 & 49.6 & 50.2\\
DCAMA (ECCV'22)~\cite{shi2022dense} & resnet50 & 41.9 & 45.1 & 44.4 & 41.7 & 43.3 & 45.9 & 50.5 & 50.7 & 46.0 & 48.3\\
BAM (CVPR'22)~\cite{kang2022integrative} & resnet50 &43.4  &50.6  &47.5  &43.4  &46.2  &49.3  &54.2  &51.6  &49.5  &51.2  \\ 
NERTNet (CVPR'22)~\cite{liu2022learning} & resnet101 & 38.3 & 40.4 & 39.5 & 38.1 & 39.1 & 42.3 & 44.4 & 44.2 & 41.7 & 43.2 \\
IPMT (NeurIPS'22)~\cite{liu2022intermediate} & resnet50 & 41.4 & 45.1 & 45.6 & 40.0 & 43.0 & 43.5 & 49.7 & 48.7 & 47.9 & 47.5 \\ 
ABCNet (CVPR'23)~\cite{wang2023rethinking} & resnet50 & 42.3 & 46.2 & 46.0 & 42.0 & 44.1 & 45.5 & 51.7 & 52.6 & 46.4 & 49.1 \\
MIANet (CVPR'23)~\cite{yang2023mianet} & resnet50 & 42.5 & 53.0 & 47.8 & 47.4 & 47.7 & 45.9 & 58.2 & 51.3 & 52.0 & 51.7 \\
MSI (ICCV'23)~\cite{moon2023msi} & resnet50 & 42.4 & 49.2 & 49.4 & 46.1 & 46.8 & 47.1 & 54.9 & 54.1 & 51.9 & 52.0 \\
\hline
PFENet (TPAMI'20)~\cite{tian2020prior} & resnet101 & 34.3 & 33.0 & 32.3 & 30.1 & 32.4 & 38.5 & 38.6 & 38.2 & 34.3 & 37.4 \\
HDMNet (CVPR'23)~\cite{peng2023hierarchical} & resnet50 & 43.8 & 55.3 & 51.6 & 49.4 & 50.0 & 50.6 & 61.6 & 55.7 & 56.0 & 56.0 \\
\hline
\textbf{ours}-PI-CLIP (PFENet) & resnet50& 36.1 & 42.3 & 37.3 & 37.7 &38.4  & 40.4 & 45.6 & 39.9 & 38.6 & 41.1  \\
\textbf{ours}-PI-CLIP (HDMNet) & resnet50& \textbf{49.3} & \textbf{65.7} & \textbf{55.8} & \textbf{56.3} & \textbf{56.8} & \textbf{56.4} & \textbf{66.2} & \textbf{55.9} & \textbf{58.0} & \textbf{59.1}  \\ \hline
\end{tabular}%
$}
\end{table*}
\begin{figure*}[h]
	\centering
	\includegraphics[width=\textwidth]{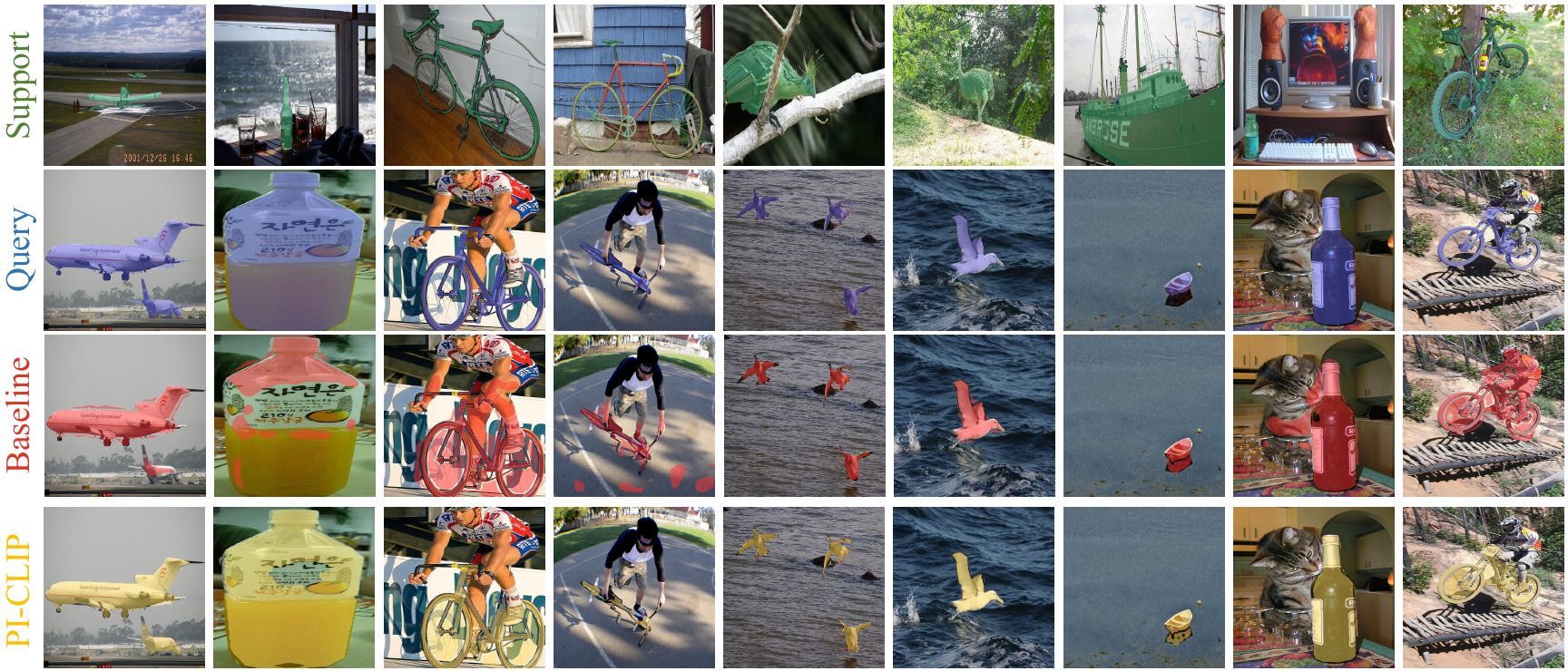}
	\caption{Qualitative results of the proposed PI-CLIP and baseline (HDMNet~\cite{peng2023hierarchical}) approach under 1-shot setting. Each row from top to bottom represents the support images with ground-truth  (GT) masks (green), query images with GT masks (blue), baseline results (red), and our results (yellow), respectively.}
	\label{fig:3}
\end{figure*}

\section{Experiments}
\textbf{Datasets and Evaluation Metrics.}
We utilize the PASCAL-5$^{i}$~\cite{shaban2017one} and COCO-20$^{i}$~\cite{nguyen2019feature} to evaluate the performance of our proposed method. PASCAL-5$^{i}$ is built on PASCAL VOC 2012~\cite{everingham2010pascal} with the complement of SDS~\cite{hariharan2011semantic} which is a classical computer vision dataset for segmentation tasks including 20 different object classes such as people, cars, cats, dogs, chairs, aeroplanes, etc. COCO-20$^{i}$ is built on MSCOCO~\cite{lin2014microsoft} consists of more than 120,000 images from 80 categories and is a more challenging dataset.
To evaluate the performance of our proposed method, we adopt mean intersection-over-union (mIoU) and foreground-background IoU (FB-IoU) as the evaluation metrics following previous works~\cite{lang2022learning, peng2023hierarchical, tian2020prior}.

\subsection{Implementation details.} We utilize HDMNet~\cite{peng2023hierarchical}, BAM~\cite{lang2022learning} and PFENet~\cite{tian2020prior} as the baseline to test our performance. In all experiments on PASCAL-5$^{i}$ and COCO-20$^{i}$, the images are set to 473$\times$473 pixels and the CLIP pre-trained model is ViT-B-16~\cite{radford2021learning}. For COCO-20$^{i}$, setting higher resolution can get higher performance but with more computing cost, the temperature parameter $\tau$ in VTP is set to 0.01 and the selected layer $l$ in PIR is set to 8. For the 5-shot case, we directly concatenate 5 VVP rather than using the average of them as the prior information. For fair comparisons, other settings like data augmentation technique, learning rate and optimizer, \eg, all follow the corresponding baselines. All experiments are run on NVIDIA V100 GPUs.
 
With the help of the accurate visual-text prior information and the generalized visual-visual prior information, our proposed PI-CLIP method can able to reach better performance quickly, so PI-CLIP is only trained for 30 epochs on both PASCAL-5$^{i}$ and COCO-20$^{i}$ which needs less time than any previous methods and the batch sizes are set to 4 on 1-shot and 2 on 5-shot respectively, the model can perform better if can be trained for more epochs.

\subsection{Comparison with state-of-the-art}
\textbf{Quantitative results.} Table~\ref{tab:tab1} shows the performance of our method and existing state-of-the-art methods for few-shot segmentation on PASCAL-5$^{i}$, our approach greatly improves the performance of the model over the 1-shot task compared to different baselines and achieves new state-of-the-art performance, with mIoU increases of 5.9$\%$ for BAM~\cite{lang2022learning} and 7.4$\%$ for HDMNet~\cite{peng2023hierarchical}. For the 5-shot segmentation task, our approach outperforms other approaches by a clear margin, with mIoU gain of 3.8$\%$ for BAM~\cite{lang2022learning} and 5.4$\%$ for HDMNet~\cite{peng2023hierarchical}, respectively. Besides, we also experimented by plugging our method into PFENet\cite{tian2020prior}, a different baseline from BAM~\cite{lang2022learning} and HDMNet~\cite{peng2023hierarchical} that does not use a base learner, it can be seen even without the inhibition of base classes by the base learner, our approach also improves the mIou of 10.4$\%$ and 10.9$\%$ for 1-shot and 5-shot tasks respectively. The performance improvement of the different baseline methods shows that our method is a plug-and-play module with high flexibility. The main reasons for our success with different approaches are the accurate localization of VTP and the strong generalization of VVP.

In Table~\ref{tab:tab2}, we compare the performance of our approach and others on COCO-20$^{i}$ dataset. Our approach also exhibits strong performance and achieves new
state-of-the-art performance. Specifically, our approach improves the baseline by 6.8$\%$ and 3.1$\%$ mIoU for 1-shot and 5-shot tasks.

\textbf{Qualitative results.}
In order to better show the effect of our proposed model on the existing methods, we visualize the results of the baseline and our proposed method in Fig.~\ref{fig:3}, it can be found that our method (yellow part) has a much stronger target localization ability than the baseline (red part), and the bias on the base class is greatly reduced.

\begin{figure*}[!t]
	\centering
	\includegraphics[width=0.95\textwidth]{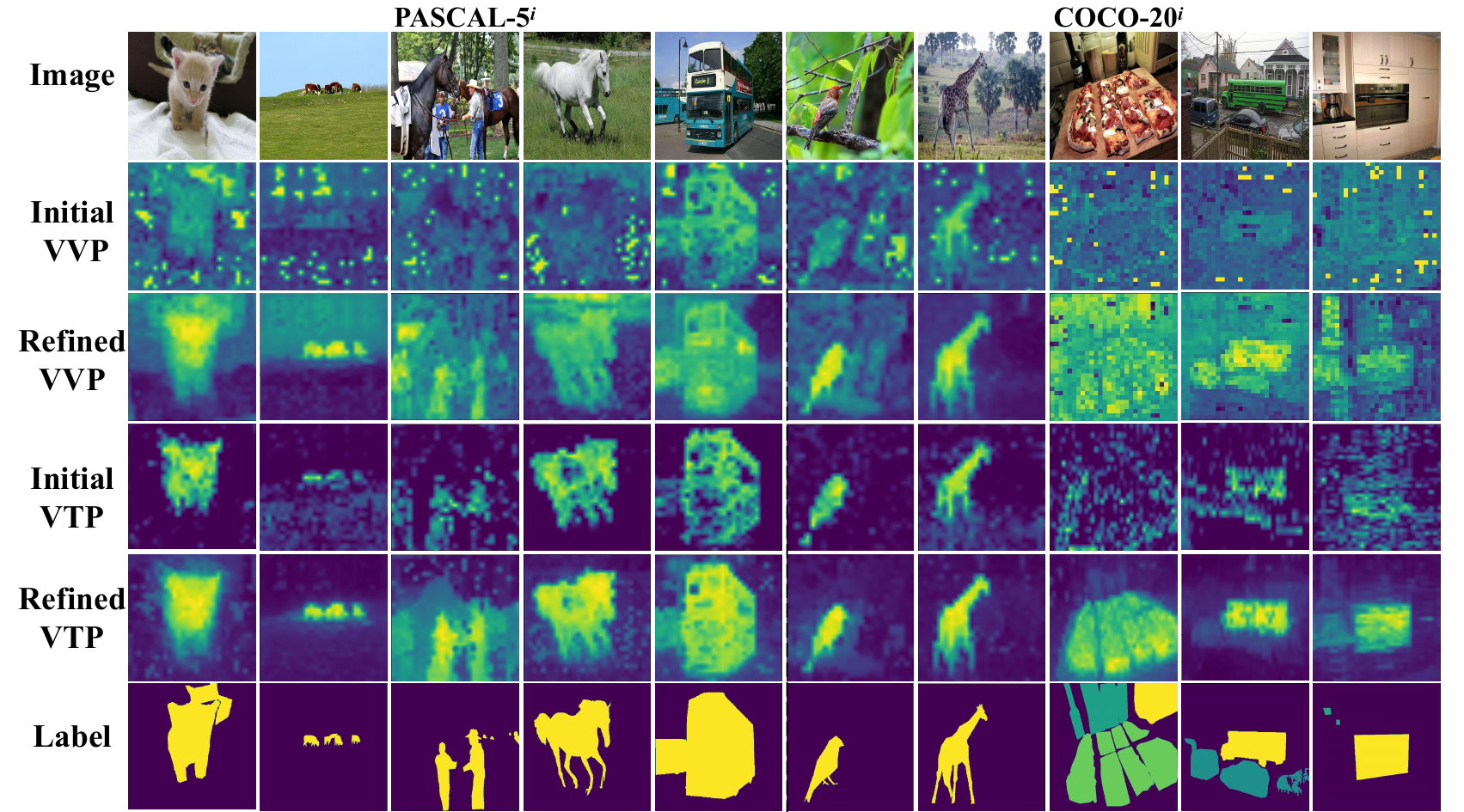}
	\caption{Visualization of the different prior information generated by our proposed method. The left is sampled from PASCAL-5$^{i}$~\cite{shaban2017one} and the right is selected from COCO-20$^{i}$~\cite{nguyen2019feature}. Each row from top to bottom represents the query image, initial visual-visual prior information, refined visual-visual prior information, initial visual-text prior information and refined visual-text prior information. The $P_{vv}$ has more general localization regions and the $P_{vt}$ has more local target regions. With the refinement of the designed high-order matrix, more accurate prior information can be extracted.}
	\label{fig:4}
\end{figure*}

Fig.~\ref{fig:4} shows the visualization of our proposed VTP and VVP to help understand the localization capabilities of VTP and the generalization capabilities of VVP. VTP focuses more on the accurate target regions, which are localized in a local region compared to the whole object. VVP, on the other hand, focuses on larger regions of the target class than VTP, but the details provided by VVP are tougher than VTP. Fig.~\ref{fig:4} also shows that synchronous refining VVP and VTP information makes them similar which is harmful to the generalization of the few-shot segmentation model.

% \zbfb{Besides, the initial information from both VVP and VTP can only provide limited details for the target, using our PIR to refine them, we can generate finer-grained guidance. By combining these two types of information, prior information with both accurate target localization ability and new class generalization ability can be obtained, which further guides the model to make more accurate segmentation.}
\begin{table}[!t]
\caption{Ablation study about our proposed VTP and VVP on the PASCAL-5$^{i}$, ``baseline" represents the HDMNet\cite{peng2023hierarchical}, VTP and VVP represent the proposed VTP module and VVP module.}
\centering
\begin{tabular}{ccccc}
\hline
baseline &VTP &VVP &mIoU ($\%$) &FB-IoU ($\%$) \\
\hline
\checkmark & & &71.00 &85.86\\
\checkmark &\checkmark & &75.30 &86.77\\
\checkmark&\checkmark &\checkmark &\textbf{76.40} &\textbf{87.57}\\
\hline
\label{tab3}
\end{tabular}
\end{table}
\subsection{Ablation Study}

We conduct a series of ablation studies to investigate the impact of each module on the PASCAL-5$^{i}$ dataset using HDMNet~\cite{peng2023hierarchical} as the baseline.

\begin{table}[!t]
\caption{Ablation study about our proposed PIR on PASCAL-5$^{i}$, $P_{vt}$ and $P_{vv}$ represent the initial information generated by VTP and VVP, PIR$_{vv}$ and PIR$_{vt}$ represent the refinement on VVP and VTP.}
\centering
\begin{tabular}{ccccccc}
\hline
$P_{vv}$ &$P_{vt}$ &PIR$_{vv}$ &PIR$_{vt}$ &mIoU ($\%$) &FB-IoU ($\%$) \\
\hline
\checkmark &\checkmark & & &75.40 &86.61\\
\checkmark &\checkmark &\checkmark & &74.82 &86.05\\
\checkmark &\checkmark & &\checkmark &\textbf{76.40} &\textbf{87.57}\\
\checkmark &\checkmark &\checkmark &\checkmark &75.70 &86.93\\
\hline
\label{tab4}
% \vspace{-3em}
\end{tabular}
\end{table}

\textbf{Ablation Study on VVP and VTP.}
The prior information has a large impact on the performance of the model, so we conduct relevant ablation studies to separately verify the validity of the prior information for the two modules we designed. As can be seen in Table~\ref{tab3}, VTP yields a performance improvement of 4.3$\%$ and VVP yields a performance improvement of 1.1$\%$.
% utilizing only the prior information from the VTP yields a performance improvement of \zbfb{x}, \zbfb{utilizing only the prior information from the VVP yields a performance improvement of x}, and the maximum performance improvement of \zbfb{x} is achieved when both are used together.
% \textbf{Ablation Study on PIR.} In the PIR module we designed a higher order matrix to maintain the structural information of the original features and used it to fine-tune the initial prior information from VVP and VTP, we conduct ablation experiments on the refinement ability of PIR as shown in Table~\ref{tab4}. It can be found that the refinement of PIR using only the VTP information as prior information is able to get an enhancement of x, and the refinement of PIR using only the VVP information as prior information is able to get an enhancement of x, and when both of them are used simultaneously, the refinement of PIR using PIR is able to achieve the highest performance of x.

\textbf{Ablation Study on PIR.} In the PIR module we designed a high order matrix to maintain the structural information of the original features and used it to refine the initial prior information, we conduct ablation experiments on the refinement ability of PIR as shown in Table~\ref{tab4}. It can be found that the refinement of PIR using only the VTP information is able to get an enhancement of 1.0$\%$, but the refinement of PIR using only the VVP information as prior information reduces model performance by 0.58$\%$, this is due to the fact that refining VVP and VTP based on the same matrix will make them produce a similar response, which will reduce the generalization of the prior guidance. When both the initial VVP and refined VTP are used, the model is able to achieve the highest performance of 76.40$\%$.

\section{Conclusion}
In this paper, we rethink the prior information for few-shot segmentation and realize that CLIP is able to achieve more accurate localization of the target class without further training. The proposed prior information generation with CLIP (PI-CLIP) can give more accurate and generalized prior information which facilitates the segmentation performance. Furthermore, we design two prior information generation modules, one is VTP which aligns the semantic information from the visual modal and text modal to generate accurate prior information, and the other is VVP which performs a matching on visual feature between support image and query image to mine more useful target information and give a regionally larger prior information. To extract more useful information, the PIR module is designed to refine the initial prior information. Extensive experiments demonstrate the effectiveness of our proposed module. In the future, we will explore how to better extract the useful information from the CLIP model.

% \paragraph{Acknowledge:} This work was supported by National Natural Science Foundation of China (No. 62301613, 62372468), the Major Basic Research Projects in Shandong Province (No.ZR2023ZD32), the Shandong Natural Science Foundation (No. ZR2023QF046, ZR2023MF008), the Taishan Scholar Program of Shandong (No. tsqn202306130), the Qingdao Natural Science Foundation (No. 23-2-1-161-zyyd-jch), Qingdao Postdoctoral Applied Research Project (No. QDBSH20230102091) and Independent Innovation Research Project of China University of Petroleum (East China) (No. 22CX06060A).
\noindent\textbf{Acknowledge:} This work was supported by National Natural Science Foundation of China (No. 62301613, 62372468), the Taishan Scholar Program of Shandong (No. tsqn202306130), the Shandong Natural Science Foundation (No. ZR2023QF046, ZR2023MF008), the Major Basic Research Projects in Shandong Province (Grant No.ZR2023ZD32), the Qingdao Natural Science Foundation (Grant No. 23-2-1-161-zyyd-jch), Qingdao Postdoctoral Applied Research Project (No. QDBSH20230102091) and Independent Innovation Research Project of China University of Petroleum (East China) (No. 22CX06060A).

{
    \small
    \bibliographystyle{ieeenat_fullname}
    \bibliography{main}

\begin{thebibliography}{64}
\providecommand{\natexlab}[1]{#1}
\providecommand{\url}[1]{\texttt{#1}}
\expandafter\ifx\csname urlstyle\endcsname\relax
  \providecommand{\doi}[1]{doi: #1}\else
  \providecommand{\doi}{doi: \begingroup \urlstyle{rm}\Url}\fi

\bibitem[Bi et~al.(2023)Bi, Feng, Yan, Mao, Diao, Wang, and Sun]{bi2023not}
Hanbo Bi, Yingchao Feng, Zhiyuan Yan, Yongqiang Mao, Wenhui Diao, Hongqi Wang, and Xian Sun.
\newblock Not just learning from others but relying on yourself: A new perspective on few-shot segmentation in remote sensing.
\newblock \emph{IEEE Transactions on Geoscience and Remote Sensing}, 2023.

\bibitem[Chen et~al.(2021)Chen, Gao, Lu, Xue, Wang, and Liao]{chen2021apanet}
Jiacheng Chen, Bin-Bin Gao, Zongqing Lu, Jing-Hao Xue, Chengjie Wang, and Qingmin Liao.
\newblock Apanet: adaptive prototypes alignment network for few-shot semantic segmentation.
\newblock \emph{arXiv preprint arXiv:2111.12263}, 2021.

\bibitem[Cheng et~al.(2021)Cheng, Schwing, and Kirillov]{cheng2021per}
Bowen Cheng, Alex Schwing, and Alexander Kirillov.
\newblock Per-pixel classification is not all you need for semantic segmentation.
\newblock \emph{Advances in Neural Information Processing Systems}, 34:\penalty0 17864--17875, 2021.

\bibitem[Cui et~al.(2023)Cui, Zhong, Tian, Liu, Yu, and Jia]{cui2023generalized}
Jiequan Cui, Zhisheng Zhong, Zhuotao Tian, Shu Liu, Bei Yu, and Jiaya Jia.
\newblock Generalized parametric contrastive learning.
\newblock \emph{IEEE Transactions on Pattern Analysis and Machine Intelligence}, 2023.

\bibitem[Deng et~al.(2009)Deng, Dong, Socher, Li, Li, and Fei-Fei]{deng2009imagenet}
Jia Deng, Wei Dong, Richard Socher, Li-Jia Li, Kai Li, and Li Fei-Fei.
\newblock Imagenet: A large-scale hierarchical image database.
\newblock In \emph{2009 IEEE conference on computer vision and pattern recognition}, pages 248--255. Ieee, 2009.

\bibitem[Ding et~al.(2023)Ding, Zhang, and Jiang]{ding2023self}
Henghui Ding, Hui Zhang, and Xudong Jiang.
\newblock Self-regularized prototypical network for few-shot semantic segmentation.
\newblock \emph{Pattern Recognition}, 133:\penalty0 109018, 2023.

\bibitem[Everingham et~al.(2010)Everingham, Van~Gool, Williams, Winn, and Zisserman]{everingham2010pascal}
Mark Everingham, Luc Van~Gool, Christopher~KI Williams, John Winn, and Andrew Zisserman.
\newblock The pascal visual object classes (voc) challenge.
\newblock \emph{International journal of computer vision}, 88:\penalty0 303--338, 2010.

\bibitem[Fan et~al.(2022)Fan, Pei, Tai, and Tang]{fan2022self}
Qi Fan, Wenjie Pei, Yu-Wing Tai, and Chi-Keung Tang.
\newblock Self-support few-shot semantic segmentation.
\newblock In \emph{European Conference on Computer Vision}, pages 701--719. Springer, 2022.

\bibitem[Guo et~al.(2023)Guo, Zhang, Qiu, Ma, Miao, He, and Cui]{guo2023calip}
Ziyu Guo, Renrui Zhang, Longtian Qiu, Xianzheng Ma, Xupeng Miao, Xuming He, and Bin Cui.
\newblock Calip: Zero-shot enhancement of clip with parameter-free attention.
\newblock In \emph{Proceedings of the AAAI Conference on Artificial Intelligence}, pages 746--754, 2023.

\bibitem[Hariharan et~al.(2011)Hariharan, Arbel{\'a}ez, Bourdev, Maji, and Malik]{hariharan2011semantic}
Bharath Hariharan, Pablo Arbel{\'a}ez, Lubomir Bourdev, Subhransu Maji, and Jitendra Malik.
\newblock Semantic contours from inverse detectors.
\newblock In \emph{2011 international conference on computer vision}, pages 991--998. IEEE, 2011.

\bibitem[He et~al.(2023)He, Jiang, Jiang, and Ding]{he2023prototype}
Shuting He, Xudong Jiang, Wei Jiang, and Henghui Ding.
\newblock Prototype adaption and projection for few-and zero-shot 3d point cloud semantic segmentation.
\newblock \emph{IEEE Transactions on Image Processing}, 2023.

\bibitem[Hong et~al.(2022)Hong, Cho, Nam, Lin, and Kim]{hong2022cost}
Sunghwan Hong, Seokju Cho, Jisu Nam, Stephen Lin, and Seungryong Kim.
\newblock Cost aggregation with 4d convolutional swin transformer for few-shot segmentation.
\newblock In \emph{Computer Vision--ECCV 2022: 17th European Conference, Tel Aviv, Israel, October 23--27, 2022, Proceedings, Part XXIX}, pages 108--126. Springer, 2022.

\bibitem[Hu et~al.(2019)Hu, Yang, Zhang, Yu, Mu, and Snoek]{hu2019attention}
Tao Hu, Pengwan Yang, Chiliang Zhang, Gang Yu, Yadong Mu, and Cees~GM Snoek.
\newblock Attention-based multi-context guiding for few-shot semantic segmentation.
\newblock In \emph{Proceedings of the AAAI conference on artificial intelligence}, pages 8441--8448, 2019.

\bibitem[Huang et~al.(2023{\natexlab{a}})Huang, Wang, Xi, and Gao]{huang2023prototypical}
Kai Huang, Feigege Wang, Ye Xi, and Yutao Gao.
\newblock Prototypical kernel learning and open-set foreground perception for generalized few-shot semantic segmentation.
\newblock In \emph{Proceedings of the IEEE/CVF International Conference on Computer Vision}, pages 19256--19265, 2023{\natexlab{a}}.

\bibitem[Huang et~al.(2023{\natexlab{b}})Huang, Dong, Yang, Huang, Lau, Ouyang, and Zuo]{huang2023clip2point}
Tianyu Huang, Bowen Dong, Yunhan Yang, Xiaoshui Huang, Rynson~WH Lau, Wanli Ouyang, and Wangmeng Zuo.
\newblock Clip2point: Transfer clip to point cloud classification with image-depth pre-training.
\newblock In \emph{Proceedings of the IEEE/CVF International Conference on Computer Vision}, pages 22157--22167, 2023{\natexlab{b}}.

\bibitem[Jiao et~al.(2023)Jiao, Wei, Wang, Zhao, and Shi]{jiao2023learning}
Siyu Jiao, Yunchao Wei, Yaowei Wang, Yao Zhao, and Humphrey Shi.
\newblock Learning mask-aware clip representations for zero-shot segmentation.
\newblock \emph{Advances in Neural Information Processing Systems}, 36:\penalty0 35631--35653, 2023.

\bibitem[Ju et~al.(2022)Ju, Zhao, Chen, Zhang, Zhang, Wang, and Tian]{ju2022adaptive}
Chen Ju, Peisen Zhao, Siheng Chen, Ya Zhang, Xiaoyun Zhang, Yanfeng Wang, and Qi Tian.
\newblock Adaptive mutual supervision for weakly-supervised temporal action localization.
\newblock \emph{IEEE Transactions on Multimedia}, 2022.

\bibitem[Kang and Cho(2022)]{kang2022integrative}
Dahyun Kang and Minsu Cho.
\newblock Integrative few-shot learning for classification and segmentation.
\newblock In \emph{Proceedings of the IEEE/CVF Conference on Computer Vision and Pattern Recognition}, pages 9979--9990, 2022.

\bibitem[Lang et~al.(2022{\natexlab{a}})Lang, Cheng, Tu, and Han]{lang2022learning}
Chunbo Lang, Gong Cheng, Binfei Tu, and Junwei Han.
\newblock Learning what not to segment: A new perspective on few-shot segmentation.
\newblock In \emph{Proceedings of the IEEE/CVF conference on computer vision and pattern recognition}, pages 8057--8067, 2022{\natexlab{a}}.

\bibitem[Lang et~al.(2022{\natexlab{b}})Lang, Tu, Cheng, and Han]{lang2022beyond}
Chunbo Lang, Binfei Tu, Gong Cheng, and Junwei Han.
\newblock Beyond the prototype: Divide-and-conquer proxies for few-shot segmentation.
\newblock \emph{arXiv preprint arXiv:2204.09903}, 2022{\natexlab{b}}.

\bibitem[Lee et~al.(2022)Lee, Yang, and Wang]{lee2022pixel}
Yuan-Hao Lee, Fu-En Yang, and Yu-Chiang~Frank Wang.
\newblock A pixel-level meta-learner for weakly supervised few-shot semantic segmentation.
\newblock In \emph{Proceedings of the IEEE/CVF Winter Conference on Applications of Computer Vision}, pages 2170--2180, 2022.

\bibitem[Li et~al.(2021)Li, Jampani, Sevilla-Lara, Sun, Kim, and Kim]{li2021adaptive}
Gen Li, Varun Jampani, Laura Sevilla-Lara, Deqing Sun, Jonghyun Kim, and Joongkyu Kim.
\newblock Adaptive prototype learning and allocation for few-shot segmentation.
\newblock In \emph{Proceedings of the IEEE/CVF conference on computer vision and pattern recognition}, pages 8334--8343, 2021.

\bibitem[Lin et~al.(2014)Lin, Maire, Belongie, Hays, Perona, Ramanan, Doll{\'a}r, and Zitnick]{lin2014microsoft}
Tsung-Yi Lin, Michael Maire, Serge Belongie, James Hays, Pietro Perona, Deva Ramanan, Piotr Doll{\'a}r, and C~Lawrence Zitnick.
\newblock Microsoft coco: Common objects in context.
\newblock In \emph{Computer Vision--ECCV 2014: 13th European Conference, Zurich, Switzerland, September 6-12, 2014, Proceedings, Part V 13}, pages 740--755. Springer, 2014.

\bibitem[Lin et~al.(2023)Lin, Chen, Wang, Wu, Li, Lin, Liu, and He]{lin2023clip}
Yuqi Lin, Minghao Chen, Wenxiao Wang, Boxi Wu, Ke Li, Binbin Lin, Haifeng Liu, and Xiaofei He.
\newblock Clip is also an efficient segmenter: A text-driven approach for weakly supervised semantic segmentation.
\newblock In \emph{Proceedings of the IEEE/CVF Conference on Computer Vision and Pattern Recognition}, pages 15305--15314, 2023.

\bibitem[Liu et~al.(2022{\natexlab{a}})Liu, Bao, Xie, Xiong, Sonke, and Gavves]{liu2022dynamic}
Jie Liu, Yanqi Bao, Guo-Sen Xie, Huan Xiong, Jan-Jakob Sonke, and Efstratios Gavves.
\newblock Dynamic prototype convolution network for few-shot semantic segmentation.
\newblock In \emph{Proceedings of the IEEE/CVF Conference on Computer Vision and Pattern Recognition}, pages 11553--11562, 2022{\natexlab{a}}.

\bibitem[Liu et~al.(2023)Liu, Nan, Zhao, Liu, Yao, Khan, Cholakkal, Anwer, Han, and Khan]{liu2023multi}
Nian Liu, Kepan Nan, Wangbo Zhao, Yuanwei Liu, Xiwen Yao, Salman Khan, Hisham Cholakkal, Rao~Muhammad Anwer, Junwei Han, and Fahad~Shahbaz Khan.
\newblock Multi-grained temporal prototype learning for few-shot video object segmentation.
\newblock In \emph{Proceedings of the IEEE/CVF International Conference on Computer Vision}, pages 18862--18871, 2023.

\bibitem[Liu et~al.(2022{\natexlab{b}})Liu, Liu, Cao, Yao, Han, and Shao]{liu2022learning}
Yuanwei Liu, Nian Liu, Qinglong Cao, Xiwen Yao, Junwei Han, and Ling Shao.
\newblock Learning non-target knowledge for few-shot semantic segmentation.
\newblock In \emph{Proceedings of the IEEE/CVF Conference on Computer Vision and Pattern Recognition}, pages 11573--11582, 2022{\natexlab{b}}.

\bibitem[Liu et~al.(2022{\natexlab{c}})Liu, Liu, Yao, and Han]{liu2022intermediate}
Yuanwei Liu, Nian Liu, Xiwen Yao, and Junwei Han.
\newblock Intermediate prototype mining transformer for few-shot semantic segmentation.
\newblock \emph{Advances in Neural Information Processing Systems}, 35:\penalty0 38020--38031, 2022{\natexlab{c}}.

\bibitem[Long et~al.(2015)Long, Shelhamer, and Darrell]{long2015fully}
Jonathan Long, Evan Shelhamer, and Trevor Darrell.
\newblock Fully convolutional networks for semantic segmentation.
\newblock In \emph{Proceedings of the IEEE conference on computer vision and pattern recognition}, pages 3431--3440, 2015.

\bibitem[Lu et~al.(2021)Lu, He, Zhu, Zhang, Song, and Xiang]{lu2021simpler}
Zhihe Lu, Sen He, Xiatian Zhu, Li Zhang, Yi-Zhe Song, and Tao Xiang.
\newblock Simpler is better: Few-shot semantic segmentation with classifier weight transformer.
\newblock In \emph{Proceedings of the IEEE/CVF International Conference on Computer Vision}, pages 8741--8750, 2021.

\bibitem[Lu et~al.(2023)Lu, He, Li, Song, and Xiang]{lu2023prediction}
Zhihe Lu, Sen He, Da Li, Yi-Zhe Song, and Tao Xiang.
\newblock Prediction calibration for generalized few-shot semantic segmentation.
\newblock \emph{IEEE Transactions on Image Processing}, 2023.

\bibitem[L{\"u}ddecke and Ecker(2022)]{luddecke2022image}
Timo L{\"u}ddecke and Alexander Ecker.
\newblock Image segmentation using text and image prompts.
\newblock In \emph{Proceedings of the IEEE/CVF Conference on Computer Vision and Pattern Recognition}, pages 7086--7096, 2022.

\bibitem[Luo et~al.(2021)Luo, Tian, Zhang, Yu, Tang, and Jia]{luo2021pfenet++}
Xiaoliu Luo, Zhuotao Tian, Taiping Zhang, Bei Yu, Yuan~Yan Tang, and Jiaya Jia.
\newblock Pfenet++: Boosting few-shot semantic segmentation with the noise-filtered context-aware prior mask.
\newblock \emph{arXiv preprint arXiv:2109.13788}, 2021.

\bibitem[Minaee et~al.(2021)Minaee, Boykov, Porikli, Plaza, Kehtarnavaz, and Terzopoulos]{minaee2021image}
Shervin Minaee, Yuri Boykov, Fatih Porikli, Antonio Plaza, Nasser Kehtarnavaz, and Demetri Terzopoulos.
\newblock Image segmentation using deep learning: A survey.
\newblock \emph{IEEE transactions on pattern analysis and machine intelligence}, 44\penalty0 (7):\penalty0 3523--3542, 2021.

\bibitem[Moon et~al.(2023)Moon, Sohn, Zhou, Yoon, Pavlovic, Khan, and Kapadia]{moon2023msi}
Seonghyeon Moon, Samuel~S Sohn, Honglu Zhou, Sejong Yoon, Vladimir Pavlovic, Muhammad~Haris Khan, and Mubbasir Kapadia.
\newblock Msi: Maximize support-set information for few-shot segmentation.
\newblock In \emph{Proceedings of the IEEE/CVF International Conference on Computer Vision}, pages 19266--19276, 2023.

\bibitem[Nguyen and Todorovic(2019)]{nguyen2019feature}
Khoi Nguyen and Sinisa Todorovic.
\newblock Feature weighting and boosting for few-shot segmentation.
\newblock In \emph{Proceedings of the IEEE/CVF International Conference on Computer Vision}, pages 622--631, 2019.

\bibitem[Okazawa(2022)]{okazawa2022interclass}
Atsuro Okazawa.
\newblock Interclass prototype relation for few-shot segmentation.
\newblock In \emph{European Conference on Computer Vision}, pages 362--378. Springer, 2022.

\bibitem[Pandey et~al.(2022)Pandey, Vardhan, Chasmai, Sur, and Lall]{pandey2022adversarially}
Prashant Pandey, Aleti Vardhan, Mustafa Chasmai, Tanuj Sur, and Brejesh Lall.
\newblock Adversarially robust prototypical few-shot segmentation with neural-odes.
\newblock In \emph{International Conference on Medical Image Computing and Computer-Assisted Intervention}, pages 77--87. Springer, 2022.

\bibitem[Peng et~al.(2023)Peng, Tian, Wu, Wang, Liu, Su, and Jia]{peng2023hierarchical}
Bohao Peng, Zhuotao Tian, Xiaoyang Wu, Chengyao Wang, Shu Liu, Jingyong Su, and Jiaya Jia.
\newblock Hierarchical dense correlation distillation for few-shot segmentation.
\newblock In \emph{Proceedings of the IEEE/CVF Conference on Computer Vision and Pattern Recognition}, pages 23641--23651, 2023.

\bibitem[Radford et~al.(2021)Radford, Kim, Hallacy, Ramesh, Goh, Agarwal, Sastry, Askell, Mishkin, Clark, et~al.]{radford2021learning}
Alec Radford, Jong~Wook Kim, Chris Hallacy, Aditya Ramesh, Gabriel Goh, Sandhini Agarwal, Girish Sastry, Amanda Askell, Pamela Mishkin, Jack Clark, et~al.
\newblock Learning transferable visual models from natural language supervision.
\newblock In \emph{International conference on machine learning}, pages 8748--8763. PMLR, 2021.

\bibitem[Rakelly et~al.(2018)Rakelly, Shelhamer, Darrell, Efros, and Levine]{rakelly2018few}
Kate Rakelly, Evan Shelhamer, Trevor Darrell, Alexei~A Efros, and Sergey Levine.
\newblock Few-shot segmentation propagation with guided networks.
\newblock \emph{arXiv preprint arXiv:1806.07373}, 2018.

\bibitem[Shaban et~al.(2017)Shaban, Bansal, Liu, Essa, and Boots]{shaban2017one}
Amirreza Shaban, Shray Bansal, Zhen Liu, Irfan Essa, and Byron Boots.
\newblock One-shot learning for semantic segmentation.
\newblock \emph{arXiv preprint arXiv:1709.03410}, 2017.

\bibitem[Shi et~al.(2022)Shi, Wei, Zhang, Lu, Ning, Chen, Ma, and Zheng]{shi2022dense}
Xinyu Shi, Dong Wei, Yu Zhang, Donghuan Lu, Munan Ning, Jiashun Chen, Kai Ma, and Yefeng Zheng.
\newblock Dense cross-query-and-support attention weighted mask aggregation for few-shot segmentation.
\newblock In \emph{Computer Vision--ECCV 2022: 17th European Conference, Tel Aviv, Israel, October 23--27, 2022, Proceedings, Part XX}, pages 151--168. Springer, 2022.

\bibitem[Shuai et~al.(2023)Shuai, Fanman, Runtong, Heqian, Hongliang, Qingbo, and Linfeng]{shuai2023visual}
Chen Shuai, Meng Fanman, Zhang Runtong, Qiu Heqian, Li Hongliang, Wu Qingbo, and Xu Linfeng.
\newblock Visual and textual prior guided mask assemble for few-shot segmentation and beyond.
\newblock \emph{arXiv preprint arXiv:2308.07539}, 2023.

\bibitem[Siam et~al.(2019)Siam, Oreshkin, and Jagersand]{siam2019adaptive}
Mennatullah Siam, Boris Oreshkin, and Martin Jagersand.
\newblock Adaptive masked proxies for few-shot segmentation.
\newblock \emph{arXiv preprint arXiv:1902.11123}, 2019.

\bibitem[Sinkhorn(1964)]{sinkhorn1964relationship}
Richard Sinkhorn.
\newblock A relationship between arbitrary positive matrices and doubly stochastic matrices.
\newblock \emph{The annals of mathematical statistics}, 35\penalty0 (2):\penalty0 876--879, 1964.

\bibitem[Sun et~al.(2023)Sun, Lu, Wang, Yin, Zhen, Snoek, and Shao]{sun2023attentional}
Haoliang Sun, Xiankai Lu, Haochen Wang, Yilong Yin, Xiantong Zhen, Cees~GM Snoek, and Ling Shao.
\newblock Attentional prototype inference for few-shot segmentation.
\newblock \emph{Pattern Recognition}, page 109726, 2023.

\bibitem[Sung et~al.(2018)Sung, Yang, Zhang, Xiang, Torr, and Hospedales]{sung2018learning}
Flood Sung, Yongxin Yang, Li Zhang, Tao Xiang, Philip~HS Torr, and Timothy~M Hospedales.
\newblock Learning to compare: Relation network for few-shot learning.
\newblock In \emph{Proceedings of the IEEE conference on computer vision and pattern recognition}, pages 1199--1208, 2018.

\bibitem[Tavera et~al.(2022)Tavera, Cermelli, Masone, and Caputo]{tavera2022pixel}
Antonio Tavera, Fabio Cermelli, Carlo Masone, and Barbara Caputo.
\newblock Pixel-by-pixel cross-domain alignment for few-shot semantic segmentation.
\newblock In \emph{Proceedings of the IEEE/CVF Winter Conference on Applications of Computer Vision}, pages 1626--1635, 2022.

\bibitem[Tian et~al.(2020)Tian, Zhao, Shu, Yang, Li, and Jia]{tian2020prior}
Zhuotao Tian, Hengshuang Zhao, Michelle Shu, Zhicheng Yang, Ruiyu Li, and Jiaya Jia.
\newblock Prior guided feature enrichment network for few-shot segmentation.
\newblock \emph{IEEE transactions on pattern analysis and machine intelligence}, 44\penalty0 (2):\penalty0 1050--1065, 2020.

\bibitem[Tian et~al.(2023)Tian, Cui, Jiang, Qi, Lai, Chen, Liu, and Jia]{tian2023learning}
Zhuotao Tian, Jiequan Cui, Li Jiang, Xiaojuan Qi, Xin Lai, Yixin Chen, Shu Liu, and Jiaya Jia.
\newblock Learning context-aware classifier for semantic segmentation.
\newblock \emph{arXiv preprint arXiv:2303.11633}, 2023.

\bibitem[Wang et~al.(2019)Wang, Liew, Zou, Zhou, and Feng]{wang2019panet}
Kaixin Wang, Jun~Hao Liew, Yingtian Zou, Daquan Zhou, and Jiashi Feng.
\newblock Panet: Few-shot image semantic segmentation with prototype alignment.
\newblock In \emph{proceedings of the IEEE/CVF international conference on computer vision}, pages 9197--9206, 2019.

\bibitem[Wang et~al.(2023)Wang, Sun, and Zhang]{wang2023rethinking}
Yuan Wang, Rui Sun, and Tianzhu Zhang.
\newblock Rethinking the correlation in few-shot segmentation: A buoys view.
\newblock In \emph{Proceedings of the IEEE/CVF Conference on Computer Vision and Pattern Recognition}, pages 7183--7192, 2023.

\bibitem[Yang et~al.(2020)Yang, Wang, Chen, Zhou, Yi, Ouyang, and Zhou]{yang2020brinet}
Xianghui Yang, Bairun Wang, Kaige Chen, Xinchi Zhou, Shuai Yi, Wanli Ouyang, and Luping Zhou.
\newblock Brinet: Towards bridging the intra-class and inter-class gaps in one-shot segmentation.
\newblock \emph{arXiv preprint arXiv:2008.06226}, 2020.

\bibitem[Yang et~al.(2023{\natexlab{a}})Yang, Chen, Feng, and Huang]{yang2023mianet}
Yong Yang, Qiong Chen, Yuan Feng, and Tianlin Huang.
\newblock Mianet: Aggregating unbiased instance and general information for few-shot semantic segmentation.
\newblock In \emph{Proceedings of the IEEE/CVF Conference on Computer Vision and Pattern Recognition}, pages 7131--7140, 2023{\natexlab{a}}.

\bibitem[Yang et~al.(2023{\natexlab{b}})Yang, Ma, Ju, Zhang, and Wang]{yang2023multi}
Yuhuan Yang, Chaofan Ma, Chen Ju, Ya Zhang, and Yanfeng Wang.
\newblock Multi-modal prototypes for open-set semantic segmentation.
\newblock \emph{arXiv preprint arXiv:2307.02003}, 2023{\natexlab{b}}.

\bibitem[Zhang et~al.(2021{\natexlab{a}})Zhang, Xiao, and Qin]{zhang2021self}
Bingfeng Zhang, Jimin Xiao, and Terry Qin.
\newblock Self-guided and cross-guided learning for few-shot segmentation.
\newblock In \emph{Proceedings of the IEEE/CVF Conference on Computer Vision and Pattern Recognition}, pages 8312--8321, 2021{\natexlab{a}}.

\bibitem[Zhang et~al.(2019)Zhang, Lin, Liu, Guo, Wu, and Yao]{zhang2019pyramid}
Chi Zhang, Guosheng Lin, Fayao Liu, Jiushuang Guo, Qingyao Wu, and Rui Yao.
\newblock Pyramid graph networks with connection attentions for region-based one-shot semantic segmentation.
\newblock In \emph{Proceedings of the IEEE/CVF International Conference on Computer Vision}, pages 9587--9595, 2019.

\bibitem[Zhang et~al.(2022{\natexlab{a}})Zhang, Navasardyan, Chen, Zhao, Wei, Shi, et~al.]{zhang2022mask}
Gengwei Zhang, Shant Navasardyan, Ling Chen, Yao Zhao, Yunchao Wei, Honghui Shi, et~al.
\newblock Mask matching transformer for few-shot segmentation.
\newblock \emph{Advances in Neural Information Processing Systems}, 35:\penalty0 823--836, 2022{\natexlab{a}}.

\bibitem[Zhang et~al.(2022{\natexlab{b}})Zhang, Shi, and Li]{zhang2022mfnet}
Miao Zhang, Miaojing Shi, and Li Li.
\newblock Mfnet: Multiclass few-shot segmentation network with pixel-wise metric learning.
\newblock \emph{IEEE Transactions on Circuits and Systems for Video Technology}, 32\penalty0 (12):\penalty0 8586--8598, 2022{\natexlab{b}}.

\bibitem[Zhang et~al.(2022{\natexlab{c}})Zhang, Zhang, Fang, Gao, Li, Dai, Qiao, and Li]{zhang2022tip}
Renrui Zhang, Wei Zhang, Rongyao Fang, Peng Gao, Kunchang Li, Jifeng Dai, Yu Qiao, and Hongsheng Li.
\newblock Tip-adapter: Training-free adaption of clip for few-shot classification.
\newblock In \emph{European Conference on Computer Vision}, pages 493--510. Springer, 2022{\natexlab{c}}.

\bibitem[Zhang et~al.(2021{\natexlab{b}})Zhang, Wei, Li, Yan, and Yang]{zhang2021rich}
Xiaolin Zhang, Yunchao Wei, Zhao Li, Chenggang Yan, and Yi Yang.
\newblock Rich embedding features for one-shot semantic segmentation.
\newblock \emph{IEEE Transactions on Neural Networks and Learning Systems}, 33\penalty0 (11):\penalty0 6484--6493, 2021{\natexlab{b}}.

\bibitem[Zhang et~al.(2023)Zhang, Gao, Jiao, Liu, and Wei]{zhang2023coinseg}
Zekang Zhang, Guangyu Gao, Jianbo Jiao, Chi~Harold Liu, and Yunchao Wei.
\newblock Coinseg: Contrast inter-and intra-class representations for incremental segmentation.
\newblock In \emph{Proceedings of the IEEE/CVF International Conference on Computer Vision}, pages 843--853, 2023.

\bibitem[Zhu et~al.(2023)Zhu, Zhang, He, Zhou, Wang, Zhao, and Gao]{zhu2023not}
Xiangyang Zhu, Renrui Zhang, Bowei He, Aojun Zhou, Dong Wang, Bin Zhao, and Peng Gao.
\newblock Not all features matter: Enhancing few-shot clip with adaptive prior refinement.
\newblock \emph{arXiv preprint arXiv:2304.01195}, 2023.

\end{thebibliography}
}

\end{document}